\documentclass[10pt,a4paper]{article}

\usepackage[utf8]{inputenc}
\usepackage[english]{babel}
\usepackage{microtype}
\usepackage[margin=0.9in]{geometry}
\usepackage{graphicx}
\usepackage{booktabs}
\usepackage{array}
\usepackage{hyperref}
\pdfmapfile{+cm.map}
\hypersetup{hidelinks,pdftitle={Page-Aware Retrieval-Augmented Generation for EvalLLM 2026},pdfauthor={Abdelhak Kelious}}
\graphicspath{{figures/}}
\title{Page-Aware Retrieval-Augmented Generation for EvalLLM 2026:\\
       A Five-Variant Study on French PDFs}
\author{Abdelhak Kelious\\Capsens, France\\\texttt{abdelhak.kelious@capsens.eu}}
\date{}

\begin{document}
\maketitle

\begin{abstract}
We study retrieval-augmented generation (RAG) for questions about French PDF documents when both the answer and its supporting document pages are evaluated. Five system variants add dense retrieval, rank fusion, reranking, and query decomposition to a BM25 baseline. On 595 challenge questions, the complete system scores 0.4450 MRR@10 and 0.4013 Recall@10, compared with 0.3430 and 0.2994 for BM25. Dense retrieval alone and a simple lexical--dense fusion both underperform BM25. Reranking improves the hybrid system, whereas adding query decomposition produces the largest further gain, with higher latency and more detected output artifacts. The complete system slightly exceeds the reported anonymous overall mean on two answer metrics but falls below it on most page-retrieval metrics. These results identify accurate page selection, rather than semantic retrieval in isolation, as the main opportunity for improvement in this setting.
\end{abstract}

\noindent\textbf{Keywords:} retrieval-augmented generation; page-level retrieval; French PDFs; BM25; reranking; query decomposition.

\section{Introduction}
Retrieval-augmented generation (RAG) conditions answers on external evidence \cite{lewis2020rag}. For PDF question answering, evidence is often evaluated at the level of individual pages: a fluent answer is insufficient when its cited page is wrong. PDF extraction noise, long documents, and names or dates that require exact matching make this a demanding retrieval task. The EvalLLM 2026 setting requires an answer accompanied by document--page references, connecting answer quality to evidence attribution \cite{petroni2020kilt,gao2023alce}.

We examine a simple progression of five systems. BM25 provides a lexical baseline \cite{robertson2009probabilistic}; dense retrieval adds semantic matching \cite{karpukhin2020dpr,wang2024multilingual}; reciprocal rank fusion (RRF) combines both rankings \cite{cormack2009rrf}; a second-stage reranker improves candidate ordering \cite{nogueira2019bert}; and query decomposition extends retrieval to multiple aspects of a question \cite{trivedi2022ircot}. The experiment asks whether each addition improves page-level evidence and generated answers, and at what runtime cost. It is a system ablation, not a claim of a new retrieval algorithm.

\section{System}
\subsection{Page-preserving indexing}
\begin{figure}[t]
  \centering
  \includegraphics[width=\linewidth]{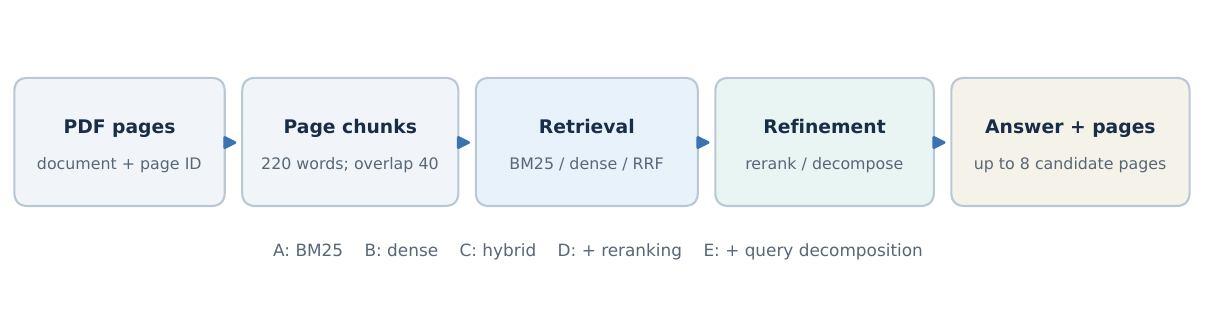}
  \caption{Page-preserving pipeline. Only the retrieval components differ across variants; answer generation selects supporting pages from the retrieved candidates.}
  \label{fig:pipeline}
\end{figure}

We extract PDF text page by page with PyMuPDF. Each page has a document name, page number, and extracted text. Long pages are divided into 220-word chunks with 40 words of overlap; every chunk retains its parent page identifier. BM25 indexes the chunks, while the dense variants embed them with \texttt{intfloat/multilingual-e5-large} \cite{wang2024multilingual}. Figure~\ref{fig:pipeline} shows the shared processing stages.

\subsection{Retrieval variants and answer generation}
Table~\ref{tab:variants} defines the runs. For hybrid retrieval, the BM25 and dense rankings are merged with RRF. For a candidate page or chunk $d$, its fused score is
\begin{equation}
 s_{\mathrm{RRF}}(d)=\sum_{r\in\{\mathrm{BM25},\mathrm{dense}\}}\frac{1}{60+\mathrm{rank}_{r}(d)},
 \label{eq:rrf}
\end{equation}
where an absent candidate contributes zero. BM25 and dense retrieval each take up to 60 candidates; the reranking variants reorder 25 candidates using \texttt{BAAI/bge-reranker-v2-m3}. Variant E also decomposes the question into short subqueries before retrieval. This description follows the reported setup; implementation details such as subquery count and the precise aggregation of chunk scores into page scores are unavailable in the supplied manuscript.

The generator receives at most eight candidate pages and is instructed to cite only pages needed to support its answer. The reported generation setting is \texttt{gpt-5.4-mini} at temperature 0.1. The original manuscript describes a shared answer prompt, generator, and context budget across variants, but does not provide the prompt text or model snapshot. Consequently, an exact reproduction of the runs would require those additional materials. Selective page citation also couples the retrieval metrics to the final citation policy: an expected page omitted at this stage cannot receive credit even if it appeared earlier among retrieved candidates.

\begin{table}[t]
\centering\small
\begin{tabular}{@{}clccc@{}}
\toprule
Run & Retrieval & RRF & Rerank & Decompose \\
\midrule
A & BM25 & -- & -- & -- \\
B & Dense & -- & -- & -- \\
C & BM25 + dense & Yes & -- & -- \\
D & BM25 + dense & Yes & Yes & -- \\
E & BM25 + dense & Yes & Yes & Yes \\
\bottomrule
\end{tabular}
\caption{Five component variants (corresponding to runs 1--5 in the supplied results).}
\label{tab:variants}
\end{table}

\section{Evaluation}
We evaluate the five runs on the 595 questions in the supplied challenge test file. Each output contains an answer and a variable-length list of cited document pages. The official report gives seven metrics. MRR@10, Recall@10, Top1, and nDCG@10 primarily measure page retrieval and ranking; LLMaaJ, QBERT, and QPARA Albert concern answer quality. Their exact scoring implementations are not specified in the supplied manuscript, so we retain the official metric names and reported scores without redefining them.

\begin{table}[t]
\centering\small
\setlength{\tabcolsep}{4pt}
\begin{tabular}{@{}lrrrrrrr@{}}
\toprule
Run & LLMaaJ & MRR & Recall & Top1 & nDCG & QBERT & QPARA \\
    &        & @10 & @10    &      & @10  &       & Albert \\
\midrule
A: BM25       & .4820 & .3430 & .2994 & .2913 & .2916 & .6935 & .7087 \\
B: dense      & .2383 & .0874 & .0703 & .0777 & .0716 & .6444 & .5534 \\
C: hybrid     & .4204 & .2168 & .2201 & .1650 & .1966 & .6826 & .6796 \\
D: reranked   & .4262 & .2573 & .2431 & .2039 & .2265 & .6708 & .7282 \\
\textbf{E: decomposed} & \textbf{.5940} & \textbf{.4450} & \textbf{.4013} & \textbf{.3592} & \textbf{.3801} & \textbf{.7093} & \textbf{.7961} \\
\midrule
Overall mean  & .5910 & .4582 & .4627 & .3789 & .4072 & .6974 & .8311 \\
Overall best  & .8416 & .7364 & .7146 & .6408 & .6570 & .7689 & .9320 \\
\bottomrule
\end{tabular}
\caption{Reported challenge scores. Bold identifies the best of our five runs in each column. The last two rows are anonymous challenge-wide summary statistics, not additional runs.}
\label{tab:scores}
\end{table}

\section{Results and analysis}
\subsection{Page retrieval and answer quality}
\begin{figure}[t]
  \centering
  \includegraphics[width=\linewidth]{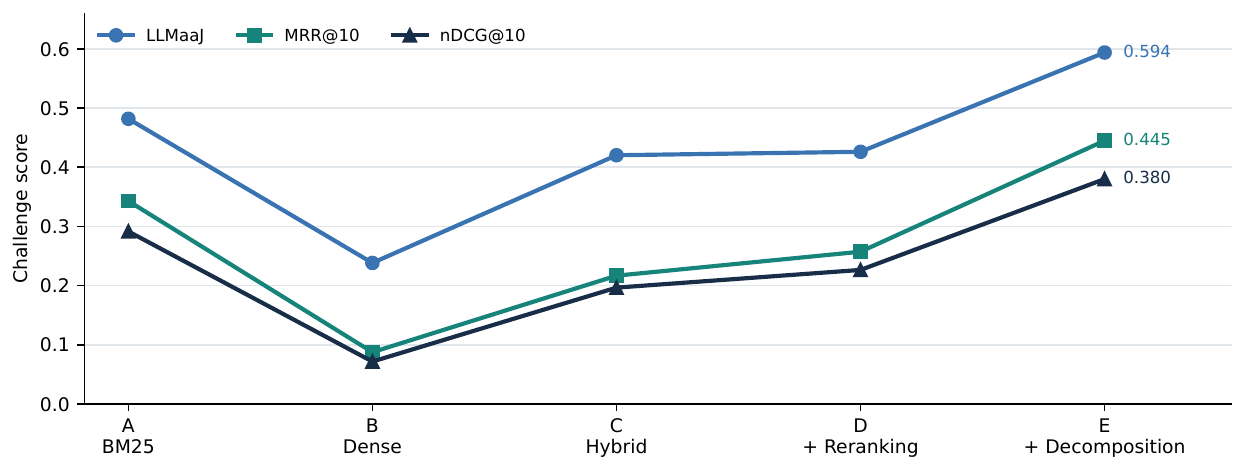}
  \caption{Reported scores for five variants. Lines connect ablation stages for readability and do not imply a continuous parameter sweep.}
  \label{fig:results}
\end{figure}

Variant E is best among our runs on all seven reported metrics (Table~\ref{tab:scores}). Compared with A, its MRR@10 rises from 0.3430 to 0.4450 (+29.7\%), Recall@10 from 0.2994 to 0.4013 (+34.0\%), and nDCG@10 from 0.2916 to 0.3801 (+30.3\%). Its QBERT increase is smaller (+2.3\%). Figure~\ref{fig:results} displays the trajectory of three representative metrics.

The additions are not uniformly beneficial. B performs substantially worse than A across all metrics, and C still trails A: for example, C reaches 0.2168 MRR@10 versus 0.3430 for BM25. A plausible explanation is that semantic similarity retrieves topically related but incorrect pages; the available aggregate results cannot establish that mechanism. D improves on C in all four page-retrieval metrics, although QBERT falls from 0.6826 to 0.6708. E produces the largest step over D: +72.9\% MRR@10, +65.1\% Recall@10, and +67.8\% nDCG@10. Without question-level results, these gains cannot specifically be attributed to long or multi-aspect questions.

Relative to the anonymous overall mean, E is slightly higher in LLMaaJ (0.5940 versus 0.5910) and QBERT (0.7093 versus 0.6974), but lower in MRR@10, Recall@10, Top1, nDCG@10, and QPARA Albert. Relative to the reported best system, its gaps are 0.2914 MRR@10, 0.3133 Recall@10, and 0.2769 nDCG@10. Page ranking and citation selection therefore remain substantial sources of error.

\subsection{Output checks}
The source manuscript reports a simple automatic screen for explicit non-answers, very short answers, unexpected writing systems, and conversational residue (Table~\ref{tab:checks}). These are counts from heuristic checks, not manually validated error rates; categories may overlap. E has the fewest non-answers (66 versus 88 for A) and very short answers (4 versus 9), but the most unexpected-script instances (21) and conversational residues (92). Better coverage thus coexists with output-control problems.

\begin{table}[t]
\centering\small
\begin{tabular}{@{}lrrrr@{}}
\toprule
Run & Non-answers & Very short & Script intrusions & Residues \\
\midrule
A & 88 & 9 & 11 & 82 \\
B & 156 & 30 & 12 & 54 \\
C & 111 & 14 & 14 & 82 \\
D & 100 & 12 & 11 & 89 \\
E & 66 & 4 & 21 & 92 \\
\bottomrule
\end{tabular}
\caption{Reported counts from automated output checks over 595 questions per run.}
\label{tab:checks}
\end{table}

\subsection{Runtime and a partial energy estimate}
Variant E takes 101.1 minutes versus 70.3 minutes for A, an increase of approximately 43.8\% (Table~\ref{tab:energy}). For context, the source manuscript estimates client energy as $E=P t$ using a constant client power of 15\,W, and associated emissions using 50\,g CO$_2$e/kWh. These assumptions do not cover remote model inference, embedding computation, or any server-side reranking; the emissions are therefore a partial client-side proxy, not a full carbon footprint. D adds only 0.8 minutes over C in these measured runs, whereas E adds 22.8 minutes over D. No variance or repeated-run measurements were supplied.

\begin{table}[t]
\centering\small
\begin{tabular}{@{}lrrr@{}}
\toprule
Run & Runtime (min) & Client energy (kWh) & Client emissions (g CO$_2$e) \\
\midrule
A & 70.3 & .01759 & .879 \\
B & 65.2 & .01630 & .815 \\
C & 77.5 & .01937 & .968 \\
D & 78.3 & .01957 & .978 \\
E & 101.1 & .02526 & 1.263 \\
\bottomrule
\end{tabular}
\caption{Reported runtime and partial client-side estimates under the stated constant-power assumptions.}
\label{tab:energy}
\end{table}

\section{Discussion and limitations}
The experiment supports a narrow but useful conclusion: reranking improves this hybrid configuration on page-ranking metrics, and the full configuration performs best among the five runs. It does not support a general claim that hybrid retrieval beats lexical retrieval, because the unfitted fusion C underperforms A. Fusion weights, page aggregation, and final citation selection merit separate analysis \cite{bruch2023hybrid}. Evaluating the candidate retrieval list before citation selection would help locate the bottleneck.

The results come from one challenge collection with no reported confidence intervals, per-question analysis, or statistical significance test. The supplied source lacks the exact prompt, decomposition procedure, and full scoring definitions. Further work should release those details; stratify questions by evidence count and type; tune fusion on held-out data; and assess citations and output artifacts manually. Such experiments would test whether the gains from decomposition reflect multi-hop evidence gathering or another change in the retrieval distribution.

Future work could also test alternative RAG approaches, such as the graph-based retrieval used in CODENS \cite{kelious2026codens}, and explore prompting strategies and how LLMs assess task-specific outputs\cite{kelious2025large,kelious2024complex}.

\section{Conclusion}
On this page-level French PDF task, BM25 is a strong baseline and dense retrieval alone is weak. Simple fusion degrades the baseline, but reranking and query decomposition recover and then improve performance. The complete system reaches 0.4450 MRR@10 and 0.4013 Recall@10 while taking longer and producing more of some detected generation artifacts. Future improvements should target page-specific ranking, citation control, and transparent accounting of inference cost.

\bibliographystyle{plain}
\bibliography{references}
\end{document}